\pdfoutput=1

\documentclass[11pt]{article}

\usepackage{naacl2021}

\usepackage{times}
\usepackage{latexsym}
\usepackage{xcolor}
\usepackage[T1]{fontenc}

\usepackage[utf8]{inputenc}
\usepackage{graphicx}
\usepackage{microtype}

\title{Can Model Compression Improve NLP Fairness?}

\author{Guangxuan Xu \\
  Department of Computer Science,\\
  University of California, Los Angeles\\
  \texttt{gxu21@cs.ucla.edu} \\\And
  Qingyuan Hu \\
  Department of Computer Science,\\
  University of California, Los Angeles\\
  \texttt{hu528@cs.ucla.edu} \\}

\begin{document}
\maketitle

\begin{abstract}
    Model compression techniques are receiving increasing attention; however, the effect of compression on model fairness is still under explored. This is the first paper to examine the effect of distillation and pruning on the toxicity and bias of generative language models. We test Knowledge Distillation and Pruning methods on the GPT2 model and found a consistent pattern of toxicity and bias reduction after model distillation; this result can be potentially interpreted by existing line of research which describes model compression as a regularization technique; our work not only serves as a reference for safe deployment of compressed models, but also extends the discussion of "compression as regularization" into the setting of neural LMs, and hints at the possibility of using compression to develop fairer models. 
\end{abstract}

\section{Introduction}
Neural Language Models, such as GPT and Bert, are fundamental not only in natural language processing (NLP) research but also in reality. Language model based applications has become an integral part of our everyday life, including AI agent Siri, Alexa, typing assistant, and Google search suggestions, etc.

However, the current neural language models come with some serious limitations - long inference time and large model size, which could limit their real-life deployment on edge devices. Model compression techniques thus becomes crucial to address such drawbacks. However, despite its importance, the  model compression's effect on fairness is still under-explored, and to the best of our knowledge, this is the first work to ever examine this issue for NLP models. 

Despite need for practical application, his topic is also intuitively interesting because there are two plausible but contradictory hypotheses regarding the effect of model compression:

(1) Memorization: Large neural models, due to over-parameterization, learns, or memorizes too well the undesired traits of the corpus they trained on. So, as the model gets compressed, it forgets some biased or toxic contents, and becomes less toxic and biased. 

(2) Winner Takes it All: Model compression reinforces heuristic bias, because as model becomes smaller, it tends to simplify the reasoning process, ignore the under-represented features, and rely on a small and highly biased subset of parameters to make predictions, making compressed models more biased.

This paper evaluates model compression's effect on toxicity and social bias. Our experiments shows that Knowledge Distillation causes monotonic reduction of model toxicity, and model bias also seem to follow a trend of reduction as model size decreases with distillation. So, the answer seems to be veering more towards the direction of the first hypothesis about memorization.

Following the intuition from previous studies that model compression could potentially improve generalization\cite{Bartoldson2020TheGT} \cite{Arani2019ImprovingGA} and robustness\cite{Goldblum2020AdversariallyRD}, we wonder if similar factors are at work here, leading to the reduction in bias and toxicity. However, this link is difficult to establish, because we don't know whether overfitting could lead to toxicity and bias increase. But this is a worthwhile question to study, and could lead to the discovery of a universal fairness improvement method for NLP; we will further investigate this connection in our next step.

Our main contributions are as follows:\\
(1) Show empirical evidence that distilled models are less toxic, and maybe less biased.

(2) Extends discussion of "model compression as a regularizer" to generative language models, and deliberate the possible applications of compression to enhance LM robustness to toxicity and bias.

\section{Related Work}

\noindent \textbf{Compression} \quad According the survey paper \cite{Gupta2020CompressionOD}, model compression methods for NLP currently include: pruning(\citet{Michel2019AreSH},\citet{Voita2019AnalyzingMS},\citet{Prasanna2020WhenBP}), quantization\cite{Cheong2019transformersZ}, knowledge distillation(\citet{Jiao2020TinyBERTDB},\citet{Iandola2020SqueezeBERTWC}), parameter sharing(\citet{Lan2020ALBERTAL},\citet{Lan2020ALBERTAL}), tensor decomposition and sub-quadratic complexity transformers. \\

\noindent \textbf{Fairness} \quad Google Brain \cite{Hooker2020CharacterisingBI} tries to characterize compression's impact on fairness for vision models. They tests quantization and pruning techniques and argue that though compressed models achieve similar overall error rate, but fairness is compromised because performance of samples with under-represented features is sacrificed after compression. Researchers from University of Utah \cite{Joseph2020GoingBC} proposes adding fairness into the compression objective function for vision tasks. However, to the best of our knowledge, no prior work has been done studying Knowledge Distillation method, nor are there any compression fairness studies on NLP models.\\

\noindent \textbf{Compression as regularization} \quad 
\cite{Fan2020ReducingTD} introduces a compression method for transformers named structured dropout, which is shown to achieve higher performance than distillation and weight pruning. The method assumes that transformer models are over-parametrized and sub-structures of the original model could achieve equivalent performances, plus that smaller networks will enjoy the benefit of regularization. Many studies (\citet{Jordo2021OnTE}, \citet{Bartoldson2020TheGT}) also argue that pruning of Convolutional Neural Networks serves as a way of regularization. \\

\noindent \textbf{Compression for robust learning} \quad
The seminal work of \cite{Papernot2016DistillationAA} introduces Knowledge Distillation as a defense against adversarial perturbations. Following works continue to use Knowledge Distillation to improve generalization \cite{Arani2019ImprovingGA} and robustness (\citet{Goldblum2020AdversariallyRD}). Knowledge Distillation is also used to improve models on privacy protection (\citet{Shejwalkar2019ReconcilingUA}, \citet{Zhao2021KnowledgeDW}). Moreover, pruning can improve model robustness according to the following studies (\citet{Jordo2021OnTE}, \citet{Pang2021BagOT},\citet{Hendrycks2019BenchmarkingNN} ). \cite{Kaya2019ShallowDeepNU} shows that stopping at earlier layers during inference can improve model robustness. The intuition is still that smaller and shallower networks are more robust.\\

\noindent \textbf{Compression for fairness} \quad Our experiments demonstrate monotonic reduction of model toxicity and biases as the model size decreases with distillation. The gold question is whether the regularization and robustness effect of model compression incur the toxicity and bias reduction that we observed in distilled generative language models. If yes, can we also develop techniques to improve NLP fairness using model compression? If not, what is the cause of the monotonic toxicity and bias reduction?

\section{Approach}

\subsection{Compression Techniques}
We conduct experiments using two methods:  Knowledge Distillation and Pruning. We choose Knowledge Distillation because it’s popular in NLP: Distill-Bert\cite{Sanh2019DistilBERTAD}, Distill-GPT, Distilled Blenderbot, and distil-T5 are all distilled generative models publicly available to download and deploy on Hugginface.co \cite{wolf-etal-2020-transformers}; those distilled models could achieve testing performance on par with original models, while cutting inference time and model size by more than half; we also test on pruning because it's the most commonly used compression techniques, and there are also works that prunes NLP models(Insert more about pruning here);\\

\noindent \textbf{Knowledge Distillation} \quad Standard training objective minimizes the cross-entropy between the model’s predicted distribution and the one-hot empirical distribution of training labels. In Distillation, rather than training with one-hot encoding, we train with the soft targets (probabilities of the teacher).  In practice, following \cite{Hinton2015DistillingTK}, we used a softmax-temperature to smoothen the soft target. A trick is to initialize student with teacher’s weight, and our experiments confirms that random initialization scores much lower performance compared with using pretrained weights. For intuition, you can view Knowledge Distillation as fine-tuning with a truncated architecture, based on soft-targets from a teacher model rather than one-hot labels. \\

\begin{figure}[ht]
\includegraphics[width=7.5cm, height=4.5cm]{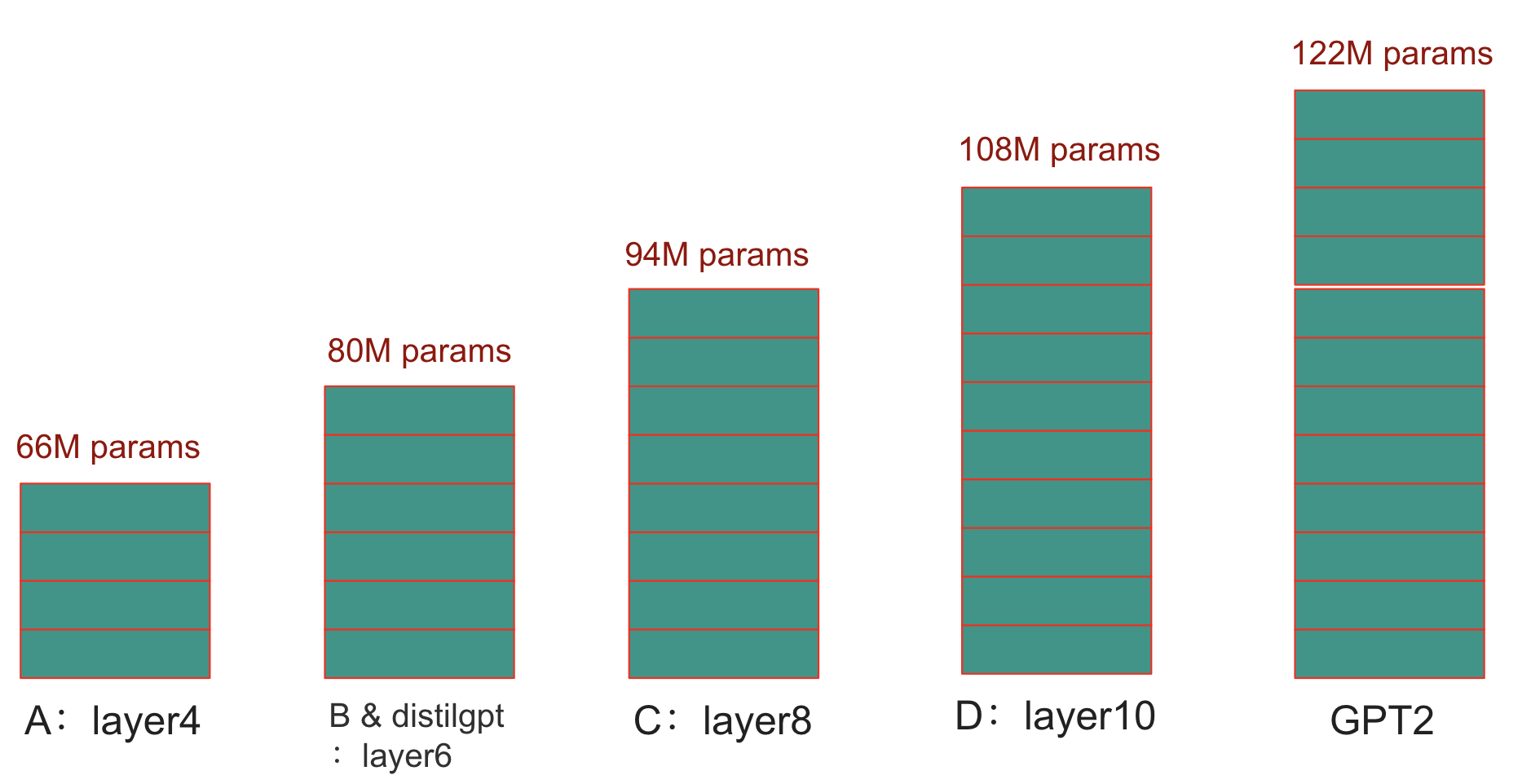}
\caption{Different classes of distilled models}
\centering
\label{fig: distill}
\end{figure}

\noindent \textbf{Pruning} \quad Pruning removes elements of a network to reduce the model size and increase inference speed. Researchers have proposed different methods to prune weights, neurons, blocks, as well as head and layers. Research shows that pruning head and layers without finetuning the models can still achieve limited performance loss. Thus, to kick off our pruning experiments, we started with pruning attention heads of GPT-2 \cite{Michel2019AreSH}. We first sort the attention heads in all layers by a proxy importance score, then mask the heads with the lowest importance score. We can also iterate this process until the loss reaches a threshold.

\subsection{Retraining Compressed Models}

\noindent \textbf{Distilled Models} \quad 
We retrained 6 distilled GPT2 language models, with different parameters sizes or training strategy. GPT-2 \cite{Radford2019LanguageMA} is a unidirectional, transformer-based model to predict the next word in a sentence. GPT2's architecture can be characterized as a stack of decoding blocks, each of which is made of self-attention mask and a fully connected 2-layer neural network.Knowledge Distillation cuts the number of decoding blocks of the model, and result in smaller models as illustrated in the Figure \ref{fig: distill}.

If we ignore the positional embedding(relatively small), the total parameters in a model is given by the sum of the word-embedding size and number-of-blocks * 7M; by this formula, the smallest A-class model is only half the size of GPT, and its inference time only 1/3 of the original GPT2.\\

\noindent \textbf{Pruned Models} \quad 
We load a pretrained GPT-2 model from HuggingFace and pruned the attention heads based on different data subsets. In our preliminary experiment, we conducted 3 iterations of pruning and kept 85 percent of the heads, namely reduced the number of heads from 144 to 122. 
In general, pruning GPT-2 models takes significant time especially on large datasets. In this paper, we report the preliminary results by varying the size of the datasets. We will continue the pruning experiment with more conditions in the future.

\subsection{Fairness Evaluation}
We run experiments to evaluate the toxicity and bias level of the pruned models; the scope of our experiments goes beyond model we retrained ourselves, but also includes some open source pre-trained models(Blenderbot\cite{Roller2021RecipesFB} and Dialogpt \cite{Zhang2020DIALOGPTL}) available on Hugginface.co. (Pruned models isn't evaluated yet due to time constraints.)\\

\noindent \textbf{Toxicity Evaluation} \quad 
For toxicity evaluation, we use the benchmark RealToxicityPrompts \cite{Gehman2020RealToxicityPromptsEN} dataset and Toxic Comment Classification Challenge(TCCC) dataset by Google Jigsaw. RealToxicityPrompts's data are sourced from the OpenWebText dataset, which is the training dataset for the original GPT2 model. The TCCC dataset is sourced from the comments pages of Wikipedia. RealToxicityPrompts contain half-sentence prompts that can be directly used to trigger Language Model generation; we curate a prompts dataset based on TCCC by cutting the number of tokens in each sentence by half. Then, we score model generations using a popular Bert-based toxicity classifier--Detoxify\cite{Detoxify}, which reported high accuracy score in Kaggle Toxic Comment Classification Challenge and Jigsaw Unintended Bias in Toxicity Classification.

\begin{figure}[ht!]
\includegraphics[scale = 0.22]{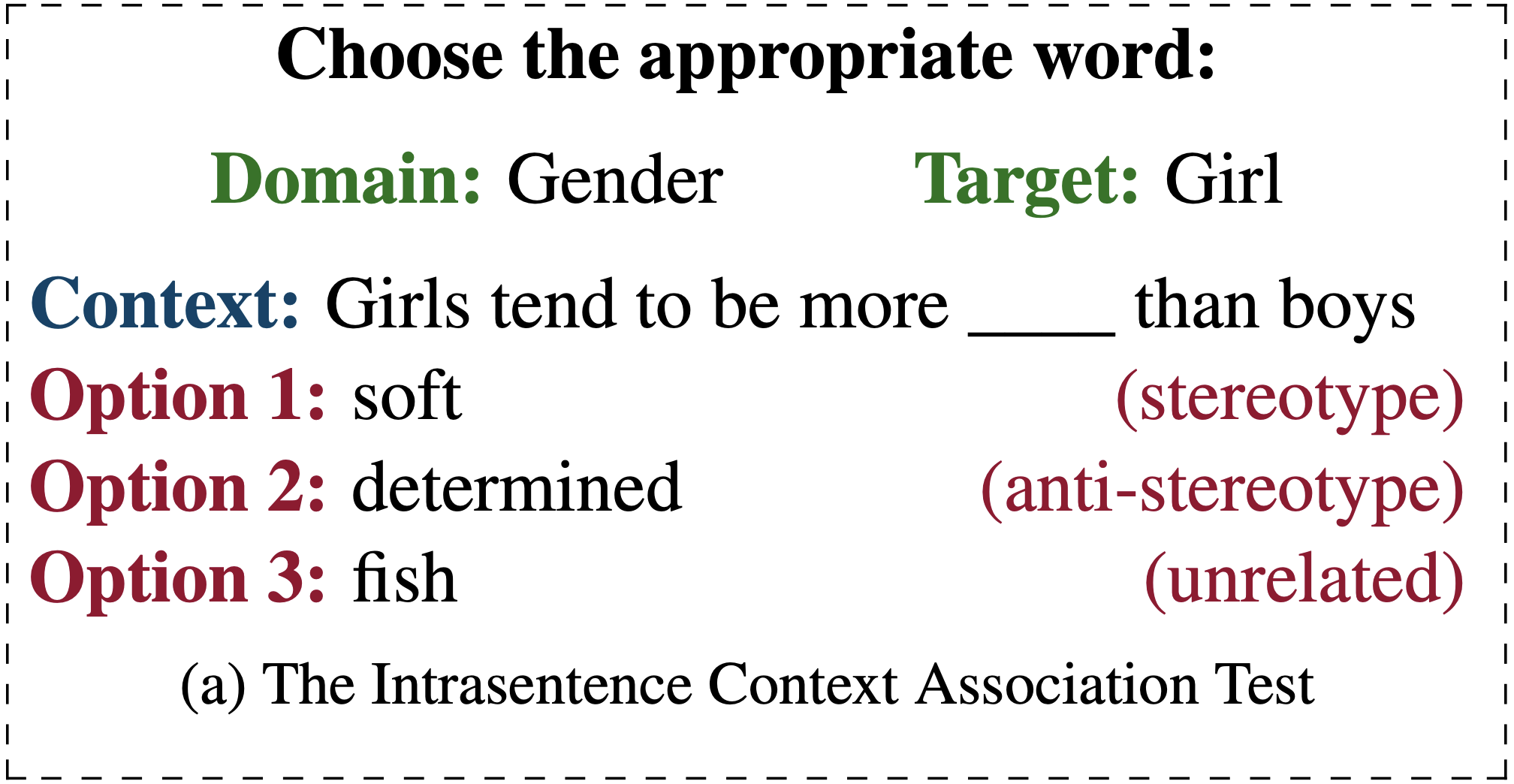}
\caption{Stereoset sample}
\centering
\label{fig: stereo}
\end{figure}

\noindent \textbf{Bias Evaluation} \quad 
For Bias Evaluation, we measure social bias using the Stereoset \cite{Nadeem2021StereoSetMS} dataset, which covers racial, national, 
gender, and professional stereotypes; The format of the dataset is given in Figure \ref{fig: stereo}; it feeds the three sentences which are either, stereotyped, anti-stereotyped, or unrelated to the LM; if the model prefers more anti-stereotyped sentences, then we could regard it as less biased. This dataset may be not as convincing and rigorous, because it can't rule out the possibility that LMs chose stereotyped sentences not for bias, but for coherence. We measure gender bias of language models using the WinoBias Dataset \cite{Zhao2018GenderBI}, which is the largest manually labelled gender bias dataset; Still, we evaluate the model by observing whether it prefers biased or anti-bias sentences, which only differ by the gender pronoun. An example of the dataset is given below:

\noindent \textbf{Anti-bias:[The farmer] asked the designer what \textcolor{orange}{[she]} could do to help. }

\noindent \textbf{Biased: [The farmer] asked the designer what \textcolor{orange}{[he]} could do to help.}

\section{Experiments}
\subsection{Model Training Details}

\begin{figure}[ht]
\includegraphics[scale = 0.4]{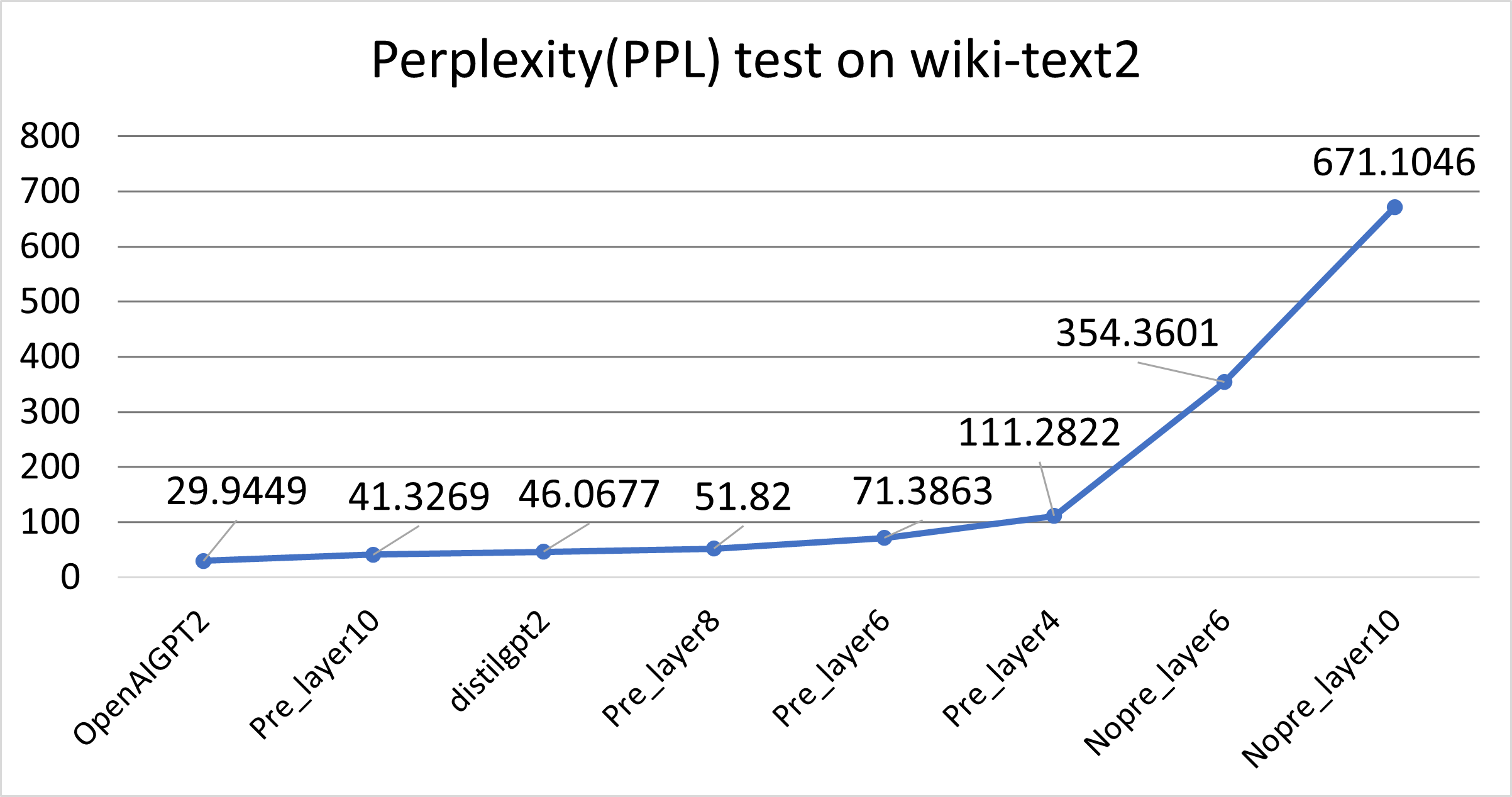}
\caption{Perplexity of distilled models}
\centering
\label{fig: ppl}
\end{figure}

\noindent \textbf{Knowledge Distillation} \quad We conduct distillation using the method described in Distill-Bert \cite{Sanh2019DistilBERTAD}. The six distilled models we trained are named (A, B, $B_{not}$, C, D, $D_{not}$), with respectively (4,6,6,8,10,10) number of decoder blocks, as shown in Figure \ref{fig: distill}; among them, $B_{not}$ and $D_{not}$ did not use pretrained weights as initialization, while all others initialized with truncated weights from the original GPT2 model. The original training dataset of OpenAI GPT2 is the the 48GB OpenWebText. However, training on full 48GB is out of our capacity, so we randomly sampled around 1/100 of the OpenWebText data to train the distilled models. Training takes 3 epochs, and it automatically saves the best performing model; The other training hyper-parameters are the same as the GPT-2 Model in huggingface. The average training time is 6 hours on 2 A100 GPUs, and the smaller the student model, the faster the training. The perplexity of trained models is shown in Figure \ref{fig: ppl}.\\

\begin{figure}[ht]
\includegraphics[scale = 0.48]{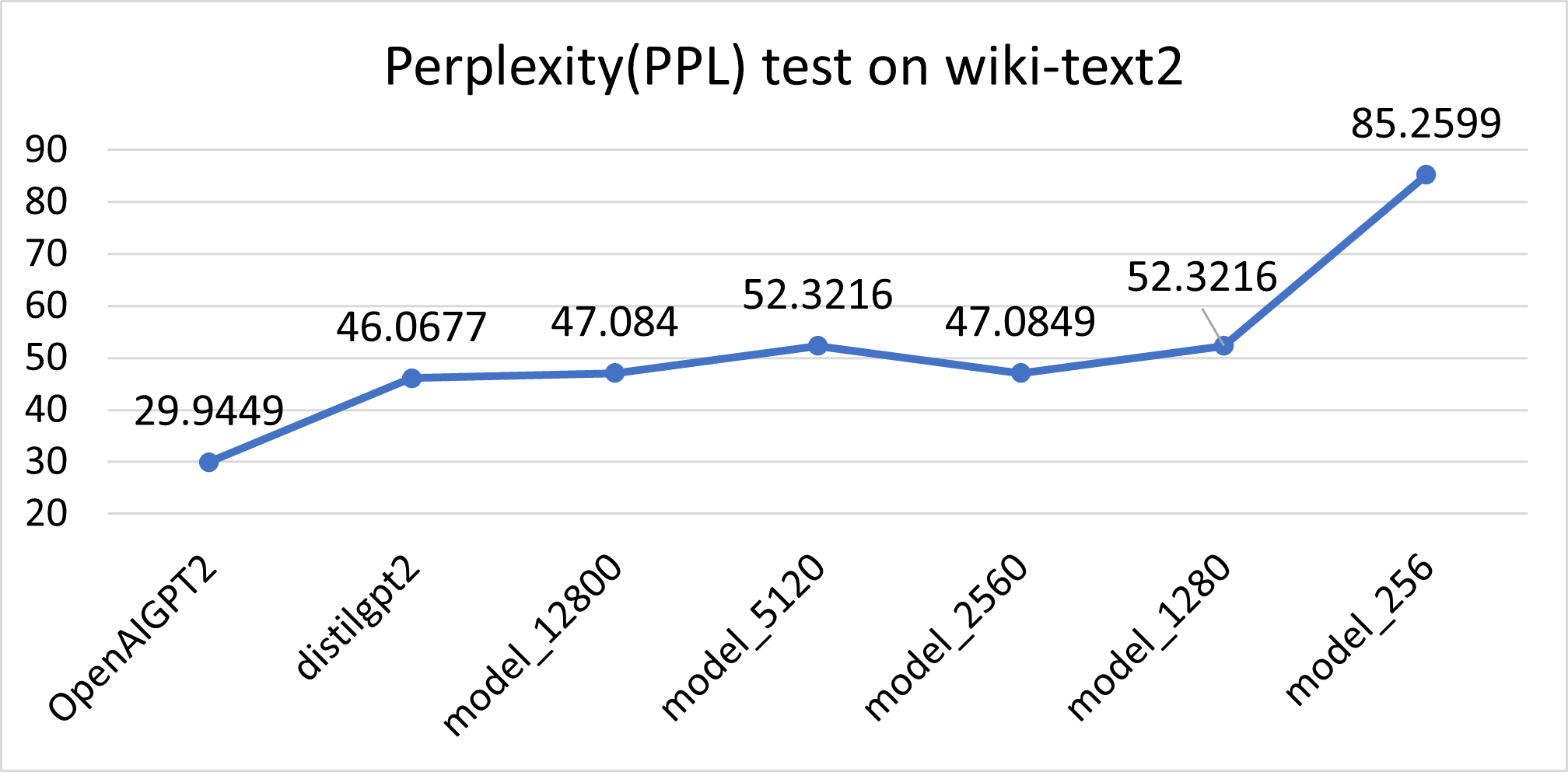}
\caption{Perplexity of pruned models}
\centering
\label{fig: prune_ppl}
\end{figure}

\noindent \textbf{Pruning} \quad Pruning applies the method proposed in \cite{Michel2019AreSH}. We start with pretrained GPT2 model from HuggingFace which has 144 attention heads in total. In each prune iteration, we sort the heads by a proxy importance score and mask the lowest important 5 percent. And we repeat the pruning 3 times which results in keeping 85 percent of the heads. We choose to use WikiText which contains a reasonable number of data (~35K). In this experiment, we sample 5 subsets with various sizes from the WikiText and conduct pruning using these subsets, namely 256, 1280, 2560, 5120, and 12800. They all have the same size, and report their perplexity (PPL) score along with the original model in Figure \ref{fig: prune_ppl}. As a result of the pruning, we observed 1.2x to 1.5x computation speed improvement. We observe that using a larger dataset as validation can help improve the pruning performance, as expected, since larger datasets results in more representative proxy importance scores.


\begin{figure*}[ht]
\includegraphics[width = \textwidth]{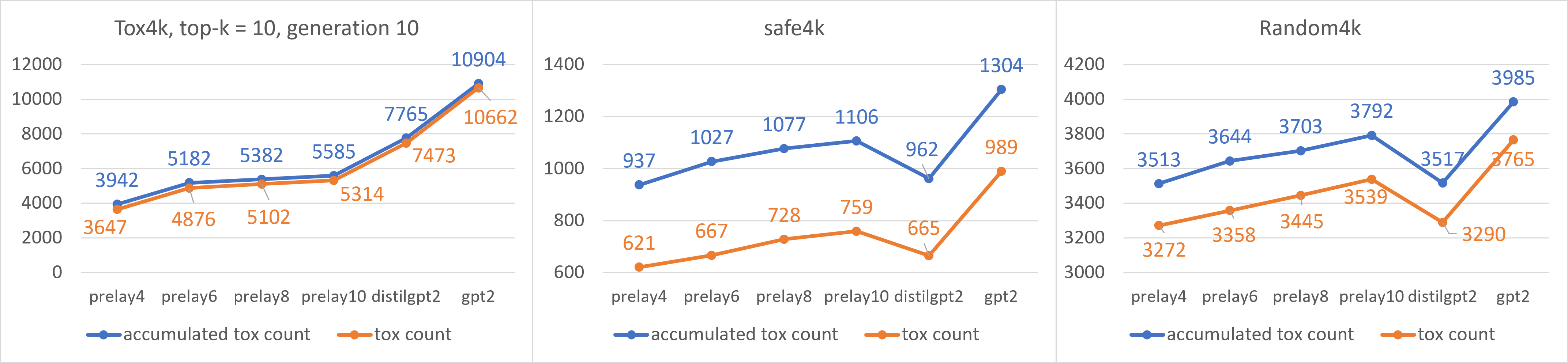}
\caption{Toxicity tested on TCCC }
\centering
\label{fig: tox_random_safe}
\end{figure*}

\subsection{Toxicity in Model Generation}
\noindent \textbf{Experiment Setup} \quad(Note that the following experiments are all conducted on Knowledge Distillation Models, because we do not have enough usable pruned models yet.) Using two prompt datasets, RealToxicityPrompts \cite{Gehman2020RealToxicityPromptsEN} and the Toxic Comments Classification Challenge(TCCC) , we run generation with different generative models and note their respective toxic count and toxic score; all models uses top-k search = 10, sample=True; as described in the previous section, we use Detoxify classifier to detect toxic examples, counting the total number of samples with toxic probability over 0.5(toxic count), and calculating the accumulated toxic probability of all samples(toxic score). For both datasets, we tried different sampling strategies, and produced the toxicity plot of different models under those settings.\\

\begin{figure}[h]
\includegraphics[scale = 0.6]{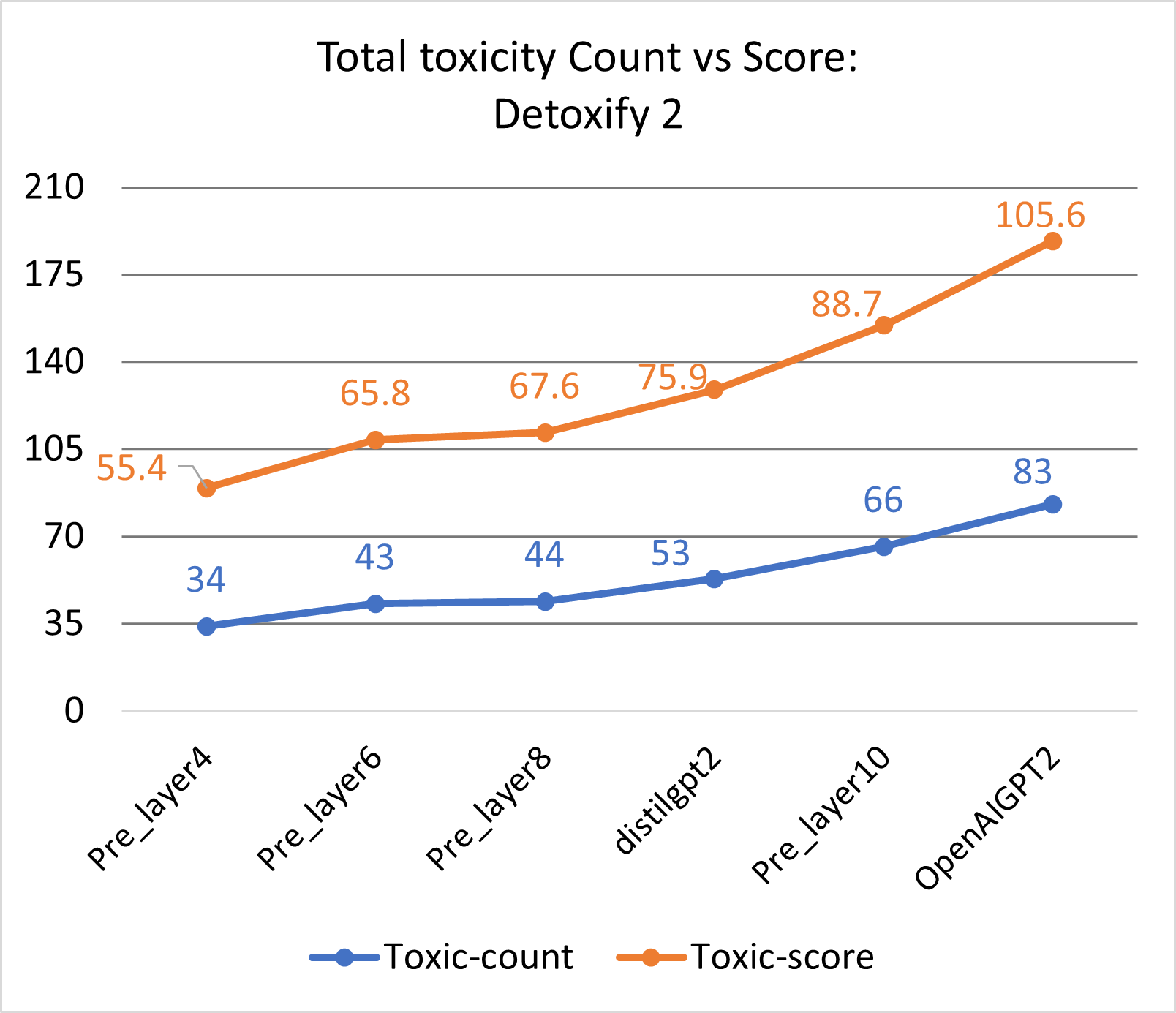}
\caption{Real Toxicity Prompts}
\centering
\label{fig: real}
\end{figure}

\noindent \textbf{RealToxicityPrompts} \quad For this dataset, we sampled 1200 toxicity triggers, which are non-toxic prompts but leads to toxic continuation in the actual data. Each model generates three sentences for each prompt, so, we are evaluating 3600 sentences per model. The toxicity level in the model generation is illustrated in Figure \ref{fig: real}. Note that RealToxicityPrompts is sourced from the training dataset of GPT2, which means the GPT2 model has seen those toxic prompts before. If the GPT2 models seems very toxic in this setting, the reason may lie in its memorization of toxic training data: distilled models are heavily regularized and less likely to output the toxic training data than the GPT2 model.\\

\noindent \textbf{TCCC Prompts} \quad Different from the RealToxicityPrompts dataset, GPT2 has never trained on the TCCC dataset. And we adopted three sampling strategies to build three test datasets, each containing 4k samples, named as(tox4k, random4k, safe4k). The tox4k contains prompts cropped from originally toxic sentences; the random4k's prompts are cropped from randomly sampled sentences in TCCC, in which around 10 percent data is toxic; the safe4k's prompts are cropped from very safe sentences, with toxic score lower than 0.001. The Figure \ref{fig: tox_random_safe} illustrate the toxicity distribution of different models under different sampling settings. To note, toxicity detection is performed on full sentence for random4k and safe4k; but only the generated part is rendered for toxicity detection in tox4k, because otherwise, all samples will be found toxic.\\

\begin{table}[hb]
\centering
\begin{tabular}{|c|c|c|c|}
\hline
            & Tox4k   & Random4k & Safe4k  \\ \hline
prelay4     & 93.7  & 110.7  & 113.3 \\ \hline
prelay6     & 91.8  & 108.4  & 110.9 \\ \hline
prelay8     & 90.4  & 107.8  & 109.6 \\ \hline
prelay10    & 89.4  & 107.6  & 110.1 \\ \hline
distillgpt2 & 165.3 & 174.9  & 175.6 \\ \hline
gpt2        & 185.2 & 197.5  & 199.0 \\ \hline
\end{tabular}
\caption{Average Length of Generation}
\centering
\label{fig: length}
\end{table}

\noindent \textbf{Observation} \quad We observe that toxicity level consistently decrease as the model size becomes smaller, in all 4 depicted sampling settings, and in two different toxicity evaluation dataset. 

\begin{figure}[ht!]
\includegraphics[scale = 0.57]{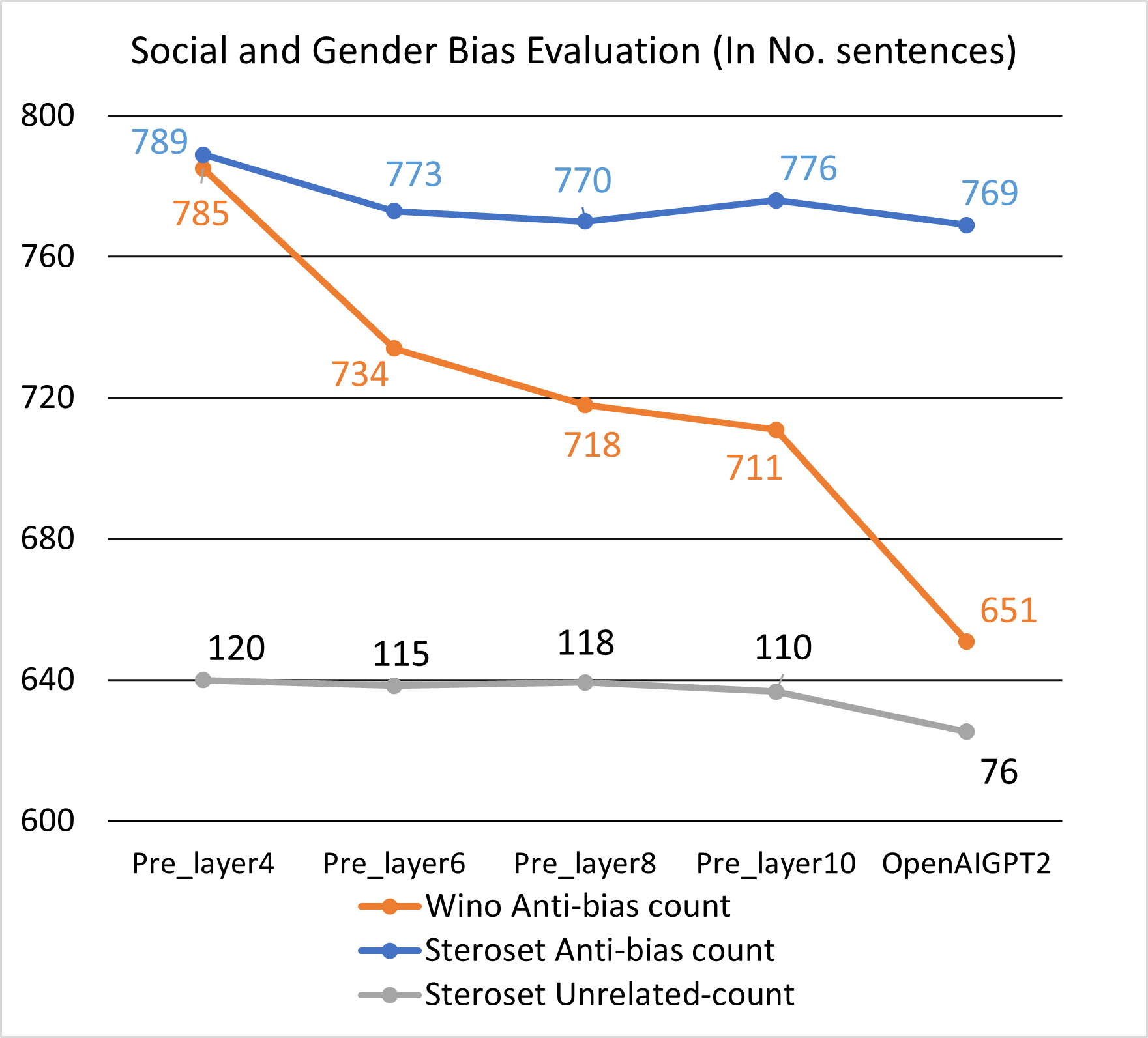}
\caption{Stereoset Bias}
\centering
\label{fig: stereobias}
\end{figure}

\subsection{Bias in Model Generation}

\noindent \textbf{Stereoset} \quad We first test the models on the Stereoset dataset, which was specifically curated by \cite{Nadeem2021StereoSetMS} to measure social bias in language models. For each model, we plot the total number of times that it chooses anti-stereotype word and unrelated words. The more anti-stereotypes it chose, the less biased it is. And the more unrelated words it choose, the more unreliable it is. The result is shown in Figure \ref{fig: stereobias}.\\

\noindent \textbf{Winobias} \quad Winobias dataset has a similar format as Stereoset, and is used to measure gender bias. It contains a total of 1584 samples, and we count the number of times each model chooses anti-bias sentences. For both datasets, the more anti-stereotypes it chose, the less biased it is. The result is shown in Figure \ref{fig: stereobias}. Other papers \cite{Barikeri2021RedditBiasAR} calculate mean perplexity difference between the anti-bias and biased sentences to measure the bias level of language models; however,since there's a significant inequality in the original perplexity of different models we compare, mean perplexity difference can be directly applied to our setting. \\

\noindent \textbf{Observation} \quad We find that gender bias level consistently decrease as the model size becomes smaller; but distillation's the effect on Stereoset dataset isn't obvious. The current experiment may be too weak to give any convincing conclusion now, but there does seem to be a trend of bias reduction after distillation.

\subsection{Ablation Study}

\begin{figure}[ht!]
\includegraphics[scale = 0.48]{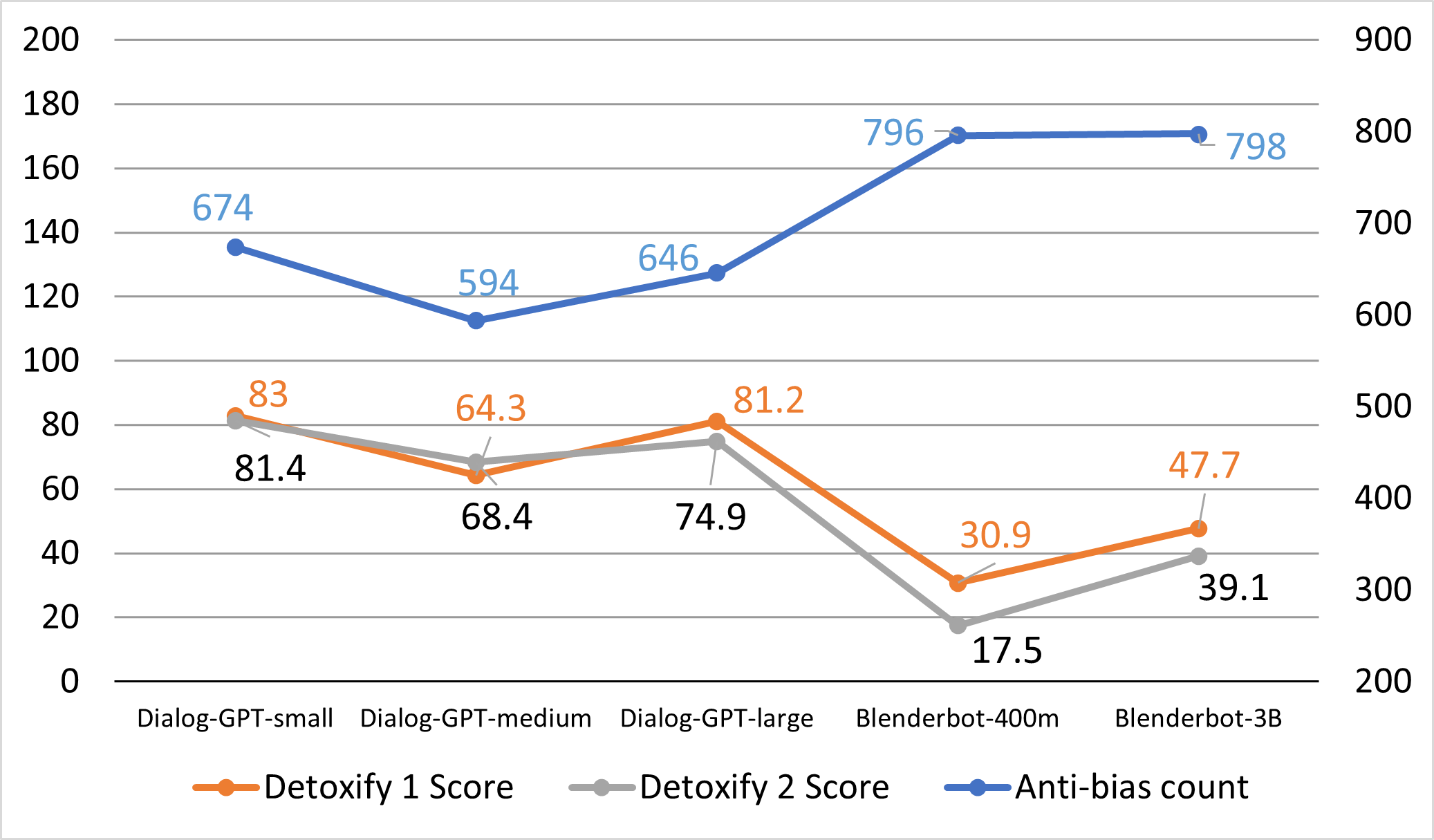}
\caption{Ablation Study}
\centering
\label{fig: Ablation}
\end{figure}

We conduct an ablation study to mitigate the concern that toxicity reduction is only the result of smaller model size. As in Figure \ref{fig: Ablation}, the DialoGPT \cite{Zhang2020DIALOGPTL} models (small, medium, large) are publicly available pretraiend models on Huggingface, trained from scratch using different architectures, rather than being distilled. The trend of toxicity and bias reduction is not observed in the DialoGPT models. Rather, toxicity decreases is observed in Distilled-400M Blenderbot versus orginal Blenderbot 3B. This study confirms that it's distillation, rather than simple model size that causes the observed fairness pattern. \\

\subsection{Perplexity vs Bias $/$ Toxicity}

\begin{figure}[ht!]
\includegraphics[scale = 0.4]{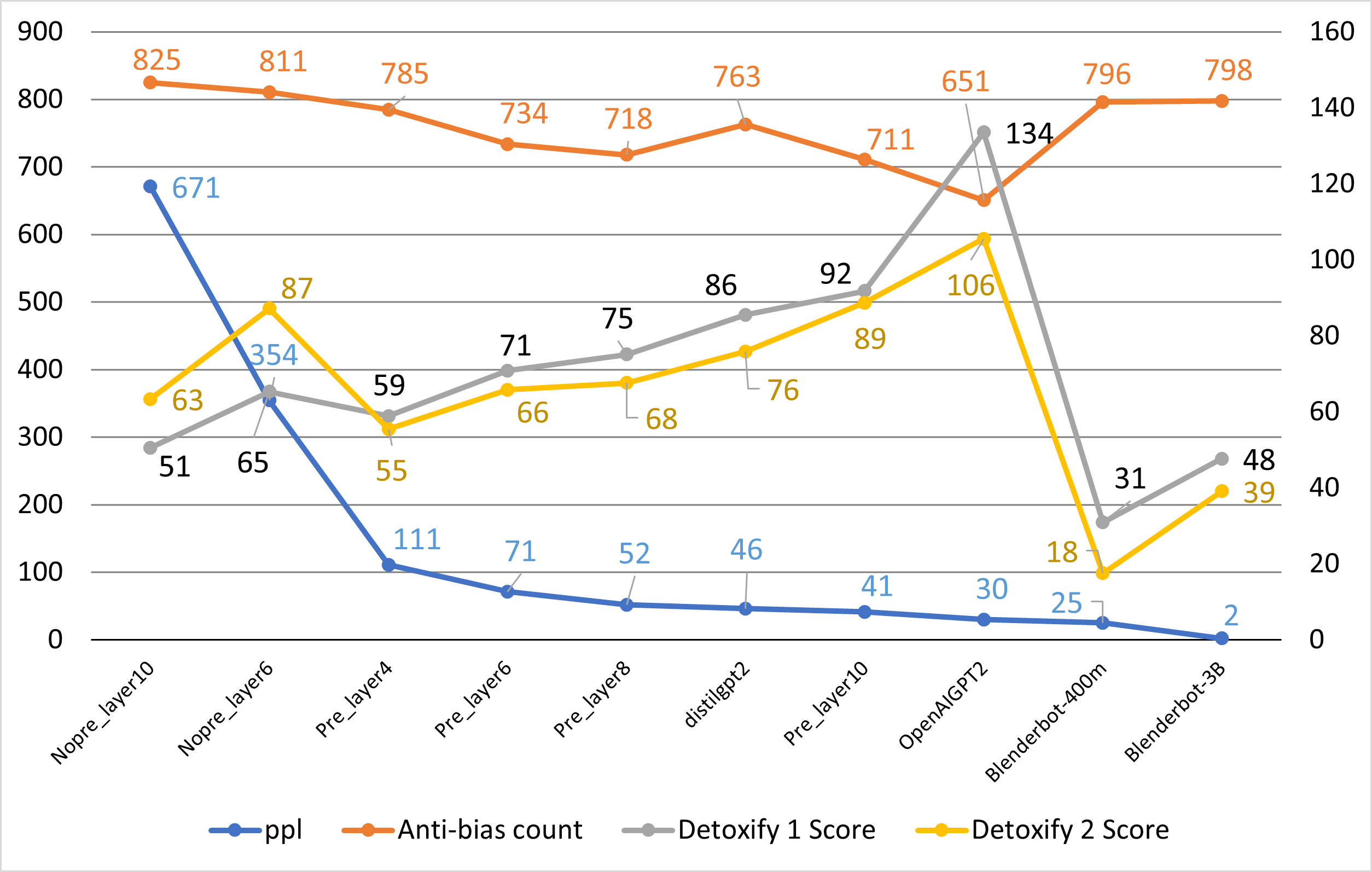}
\caption{Perplexity vs Bias $/$ Toxicity}
\centering
\label{fig: ppl_bias_t}
\end{figure}

Figure \ref{fig: ppl_bias_t} illustrates a potential weakness of our experiments; that is as the PPL decreases, model toxicity and bias level increases. Then, it can be argued that it's low perplexity that caused models to seem less biased and less toxic. However, in general, it's difficult to match heavily compressed model's PPL with the original model, so this concern couldn't be easily addressed. What we can do in the future is to narrow down the perplexity differences, and observe if there's any decrease in bias/toxicity reduction. 

\section{Discussion and Conclusion}

\noindent \textbf{Result for Toxicity Evaluation} \quad In order to make the result exhaustive and convincing, we conducted experiments using two different datasets, with four different sampling strategies. The result is surprisingly coherent and uniform, that toxicity decreases with the intensity of distillation. We first experiment with toxic triggers from RealToxicityPrompts, which is sourced from the training dataset of GPT2. If the full GPT2 model produces more toxicity on this dataset, it could be because that full GPT2 memorizes more toxic training data; this hypothesis is supported by the fact that GPT2 has much longer generation length than the distilled models as in Figure \ref{fig: length}, implying that it remembers more information regard this given prompts. However, if this is the case, the toxicity reduction pattern may not persist under another prompt sampling setting;\\

So, we added experiments using the TCCC datasets, which the GPT2 has never seen. We also experimented with three sampling strategy: toxic, safe, and random. The result is shown in Figure \ref{fig: tox_random_safe}, in which the same toxicity reduction pattern does persist. This result suggests that a simple memorization explanation can't account for this phenomenon, since no models has seen the TCCC data before. We are unable to explain this result now, and will try to give interpretation in the future.\\ 

\noindent \textbf{Result for Bias Evaluation} \quad The experiment for bias reduction is recorded in Figure \ref{fig: stereobias}, but is still short of being conclusive. Gender bias as measured by Winobias dataset clearly decreases with distillation. However, the social bias as measured by the Stereoset have no such pattern and largely stays flat. I think there are three main possibilities. The first one is that any pattern recorded is due to randomness, since the testing dataset is relatively small(1.5k). The second possibility is that Stereoset did not give good reflection of model bias, because models make choices not based on bias, but baed on coherence. Stereoset is not a very rigorous benchmark for bias. The third possibility is that Winobias did not give good measurement of bias, which is quite unlikely, since the only variable in Winobias is the gender pronoun. Winobias results should be much more reliable than that of the Stereoset. \\

\noindent \textbf{In conclusion,} \quad this paper evaluates the effect of model compression techniques (knowledge distillation and pruning) on NLP fairness. The pattern that distillation improves model fairness is being recorded, but we could not very well explain this phenomenon. We do hypothesize that this effect could be connected to regularization and model robustness, but still needs further experiments and theoretical support to verify this connection.


\section{Future Work}

\noindent \textbf{Fairness Evaluation} \quad We will add more experiment to verify our observation about bias reduction. The current evaluation dataset is too small, and the method may also be a bit naive. We would consider using bias regard metric \cite{Sheng2019TheWW} and try measuring bias in word embedding;\\

\noindent \textbf{Compression} \quad We will investigate the effectiveness of the current technique of pruning attention heads, and maybe try weights pruning and structured dropout. The toxicity and bias experiments for pruning methods will also be added. Moreover, the training time has become a large constraint given the large GPT2 model size; we will try to design experiments on smaller architectures in the future.\\

 \noindent \textbf{Theoretical explanation} \quad We have thus far unable to verify our hypothesis about the connection between regularization and the observed fairness improvement. However, we believe that if Knowledge Distillation and Pruning can effectively improve model robustness against adversarial attacks\cite{Papernot2016DistillationAA}, it is highly likely that they can also improve LM fairness. The prospect of developing universal fairness techniques for LMs based on compression is very promising. 

\newpage
\bibliography{anthology,custom}
\bibliographystyle{acl_natbib}

\appendix


\end{document}